\documentclass[journal]{IEEEtran}
\IEEEoverridecommandlockouts
\usepackage{orcidlink}
\usepackage{cite}
\usepackage{amsmath,amssymb,amsfonts}
\usepackage{algorithmic}
\usepackage{graphicx}
\usepackage{textcomp}
\usepackage{xcolor}
\usepackage{subfig}  

\def\BibTeX{{\rm B\kern-.05em{\sc i\kern-.025em b}\kern-.08em
    T\kern-.1667em\lower.7ex\hbox{E}\kern-.125emX}}
\makeatletter
\def\fps@figure{tbp}
\def\fps@figure*{tbp}
\makeatother
\begin{document}

\title{Seeing Structural Failure Before it Happens: An Image-Based Physics-Informed Neural Network (PINN) for Spaghetti Bridge Load Prediction\\
}

\author{%
    Omer Jauhar Khan\orcidlink{0009-0006-4203-4665},
    Hafeez Anwar\orcidlink{0000-0001-9529-3966},
    Sudais Khan\orcidlink{0009-0001-6473-6100},  
    Shahzeb Khan\orcidlink{0009-0007-6214-2046},
    Shams Ul Arifeen\orcidlink{0000-0002-9222-7610}
    and Farman Ullah\orcidlink{0000-0002-2488-8353}
    \thanks{Omer Jauhar Khan, Hafeez Anwar, Sudais Khan, Shahzeb Khan, and Shams Ul Arifeen are with the Department of Computer Science, National University of Computer and Emerging Sciences (FAST-NUCES), Peshawar, 25000, Pakistan (email: p218055@pwr.nu.edu.pk, hafeez.anwar@nu.edu.pk, p218033@pwr.nu.edu.pk, shahzeb.khan@nu.edu.pk, shams.arifeen@nu.edu.pk).}
    \thanks{Farman Ullah Gul Rauf is an Assistant Professor with the Department of Computer and Network Engineering, United Arab Emirates University (UAEU), Al Ain, United Arab Emirates (email: Farman@uaeu.ac.ae).}
}

\maketitle

\begin{abstract}
Physics Informed Neural Networks (PINNs) are gaining attention for their ability to embed physical laws into deep learning models, which is particularly useful in structural engineering tasks with limited data. This paper aims to explore the use of PINNs to predict the weight of small scale spaghetti bridges, a task relevant to understanding load limits and potential failure modes in simplified structural models. Our proposed framework incorporates physics-based constraints to the prediction model for improved performance. In addition to standard PINNs, we introduce a novel architecture named Physics Informed Kolmogorov Arnold Network (PIKAN), which blends universal function approximation theory with physical insights. The structural parameters provided as input to the model are collected either manually or through computer vision methods. Our dataset includes 15 real bridges, augmented to 100 samples, and our best model achieves an $R^2$ score of 0.9603 and a mean absolute error (MAE) of 10.50 units. From applied perspective, we also provide a web based interface for parameter entry and prediction. These results show that PINNs can offer reliable estimates of structural weight, even with limited data, and may help inform early stage failure analysis in lightweight bridge designs. \newline
The complete data and code are available at \url{https://github.com/OmerJauhar/PINNS-For-Spaghetti-Bridges}.

\end{abstract}

\begin{IEEEkeywords}
physics-informed neural networks, Kolmogorov-Arnold networks, structural analysis, parameter extraction, spaghetti bridges, structural mechanics
\end{IEEEkeywords}
\section{Introduction}
Physics Informed Neural Networks (PINNs) have emerged as a powerful class of models that integrate known physical laws into the training process of neural networks~\cite{raissi2019physics}. By embedding physics directly into the loss function, PINNs ensure that the learned solutions not only fit the data but also satisfy governing physical relationships. This hybrid approach has demonstrated superior generalization and robustness compared to purely data-driven methods, particularly in settings with limited or noisy data. PINNs have been successfully applied in a range of scientific and engineering domains, including fluid dynamics~\cite{ren2024physics}, biomedical engineering~\cite{kissas2020machine}, geophysics~\cite{secci2024physics}, and heat transfer problems~\cite{jalili2024physics}, where they have consistently outperformed conventional neural networks in producing physically consistent and interpretable results.

In structural engineering, traditional analysis techniques such as those based on finite element methods, have long provided reliable results, but often at the cost of high computational expense and a reliance on detailed, sometimes hard-to-measure input parameters~\cite{liu2025investigation, dilsiz2025field}. As structural systems grow in complexity and scale, there is a growing need for predictive models that are not only computationally efficient but also capable of generalizing across varying configurations~\cite{zhang2025full, das2025hybrid}. While machine learning (ML) methods offer promising alternatives, their purely data-driven nature can lead to predictions that overlook essential physical constraints and may be physically implausible~\cite{urban2025unveiling}. In this context, PINNs offer a compelling middle ground that blends data-driven flexibility with embedded domain knowledge to produce results that are both accurate and physically meaningful~\cite{chew2025physics, wang2025physics}.

To explore the practical application of PINNs within the context of structural engineering, we focus on spaghetti bridges which is a simplified yet instructive case. These small scale models serve as effective proxies for investigating the behavior of larger structural systems due to which they are widely used in educational settings to demonstrate fundamental principles of structural mechanics~\cite{groth2018framework}. Despite their simplicity, spaghetti bridges exhibit key mechanical behaviors found in full scale structures, including load transfer, stress distribution, and characteristic failure modes~\cite{mahendran1995project}.

In this work, we develop a physics-informed learning framework that predicts the load of spaghetti bridges using structural parameters obtained either manually or through computer vision applied to bridge images. Beyond its educational utility that allows students to evaluate their designs prior to physical testing, our framework demonstrates how integrating domain knowledge with machine learning can significantly improve prediction accuracy particularly in data-constrained settings. At the core of our approach is a novel hybrid neural architecture, the Physics Informed Kolmogorov Arnold Network (PIKAN) that merges universal function approximation theory with physical constraints derived from structural behavior. Despite working with a limited dataset of 100 physical bridges, our model delivers strong predictive performance, highlighting the potential of PINNs for lightweight structural modeling. We further develop a web-based interface for interactive parameter input and prediction, aiming to bridge the gap between theory and hands-on learning.

To summarize, this paper introduces the following key contributions:
\begin{itemize}
    \item A novel Physics Informed Kolmogorov Arnold Network (PIKAN) that predicts the weight of spaghetti bridges while embedding physical knowledge, in contrast to traditional black-box machine learning models.
    \item An image-based pipeline for detecting and extracting structural parameters from spaghetti bridge photographs.
    \item A curated dataset comprising 15 physical spaghetti bridges, augmented to 100 samples for training and evaluation.
    \item A user-friendly, web-based interface that accepts bridge parameters and returns weight predictions, designed as an accessible tool for students and educators.
\end{itemize}

The rest of the paper is organized as follows: Section 2 reviews related work in structural analysis using machine learning. Section 3 provides a system overview. Section 4 details our parameter extraction approaches. Section 5 explains our dataset preparation. Section 6 describes the neural network architecture with physics-informed enhancements. Section 7 presents our novel PIKAN architecture. Section 8 presents experimental results, followed by discussion in Section 9. Section 10 outlines future work, and Section 11 concludes the paper.
\section{Related Work}
Our main aim is to predict the maximum stress load of a spaghetti bridge by utilizing their physical parameters in the training process of a PINN and PIKANN. To this end, in this section, we briefly give the overview of some of the most recent works that have used PINN for load or response prediction of various types of bridges under various load conditions.

A physics-informed neural network (PINN) framework is proposed to predict the dynamic responses of bridges under moving loads~\cite{li2025moving}. The partial differential equations and boundary conditions are integrated into the training process of the model that achieves accurate spatiotemporal stress and displacement predictions. Consequently, the approach reduces reliance on dense sensor networks and shows high accuracy across various load scenarios.

A physics-informed neural network (PINN)-based framework is propsed to identify bridge influence lines and dynamic loads from multiple vehicles~\cite{li2025identification}. Using sparse sensors data, their model integrates bridge mechanics into the training process of PINN to predict structural responses. The deployment of PINNs achieve improved results by accurately estimating both the location and magnitude of loads compared to the conventional ML-based approaches.

The PINN-based structural health monitoring is performed by modeling the behavior of Kirchhoff–Love plates that are used to represent bridge decks and thin-walled structures~\cite{al2024physics}. With limited and noisy data, the plate theory-induced learning process of PINN results in the successful detection of structural damage. Consequential, the achieved results demonstrate that PINNs can effectively capture physical behavior and damage locations without relying on conventional dense instrumentation.

a physics-informed recurrent neural network (PI-RNN) is proposed for the reconstruction of dynamic displacement in bridges under moving loads~\cite{tao2025dynamic}. The RNN architecture is enriched with the physical properties of the bridge by integrating their respective equations. As a result, the PI-RNN model accurately recovers full-field displacements from sparse acceleration data. The proposed method is shown to outperform conventional neural networks demonstrating strong potential for real-time structural health monitoring.

Yet again, under moving loads on girder bridges, a PINN-based virtual sensing method is proposed to predict structural responses~\cite{al2025developing}. By embedding physical laws into the training process, the proposed model successfully reconstructs displacements and internal forces using limited sensor data. The approach significantly reduces dependence on dense instrumentation and provides accurate, real-time estimations for bridge monitoring.

For predicting the response of highway bridges under dynamic loads due to moving vehicles, a PINN-based approach is proposed~\cite{yousefpour2025physics}. By incorporating governing equations into the learning framework, the model accurately simulates displacement and acceleration with minimal sensor input. The results demonstrate that PINNs can offer a reliable, physics-consistent alternative to conventional numerical simulations in bridge engineering.

A novel PINN-based framework is proposed for identifying damage in multi-degree-of-freedom reinforced concrete bridge piers subjected to nonlinear dynamic loading~\cite{yamaguchi2024physics}. The model integrates nonlinear structural dynamics into the training process of PINN, enabling detection of damage parameters without requiring full-scale physical sensors.

Most of these studies focus on large structures and require detailed information or many sensors. One related study by Patel and Rodriguez~\cite{b4} used traditional neural networks to predict the failure load of small-scale educational bridges. Although their results are encouraging, their approach relied on manual geometric measurements and did not include any physics-based constraints in the learning process. In our work, we take a different path by introducing physics informed learning to spaghetti bridge modeling. To the best of our knowledge, this is the first study that applies a physics informed neural network to these types of bridges. We believe this approach offers better reliability because it not only learns from the data but also respects the underlying physical rules of structural behavior. To achieve this, we propose a new model called the Physics Informed Kolmogorov Arnold Network, which combines mathematical learning with structural physics. Inspired by StrucNet~\cite{b2}, we also present a computer vision method that detects and extracts bridge parameters directly from images, making the system more practical and easier to use. This combination allows us to show that accurate predictions are possible even when the dataset is small, which is often the case in educational setups.

\section{Dataset Preparation}

\begin{figure}[!t]
\centering
\includegraphics[width=\columnwidth]{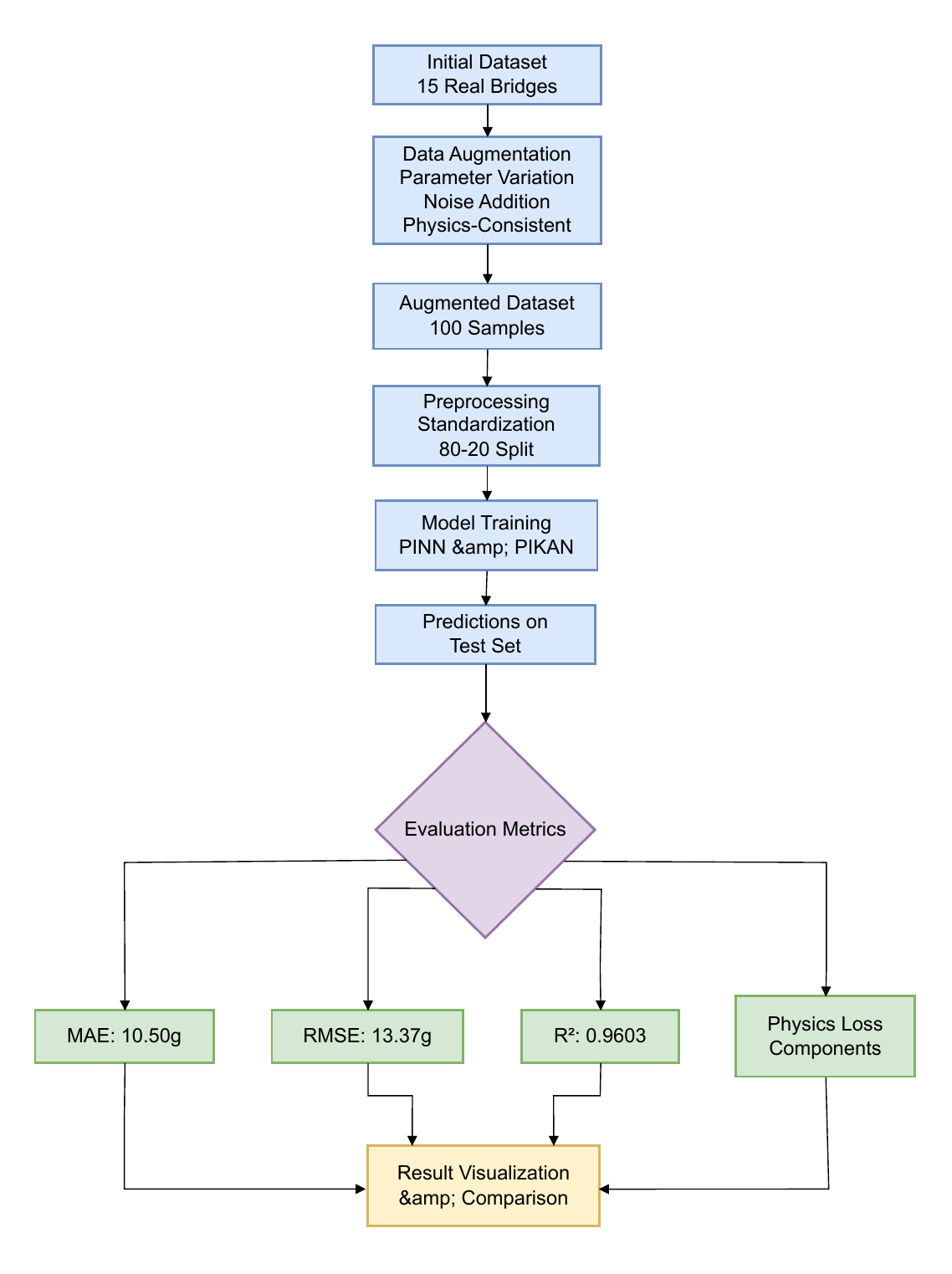}
\caption{Overview of the complete methodology pipeline:  (1) Initial dataset of 15 real bridges with measured geometric and material properties, (2) Data augmentation through parameter variation, noise addition, and physics-consistent transformations to generate 100 samples, (3) Preprocessing including standardization and 80-20 train-test split, (4) Training of PINN and PIKAN models with physics-informed loss functions, (5) Predictions on test set, (6) Evaluation metrics achieving MAE of 10.50g, RMSE of 13.37g, and R² of 0.9603, (7) Result visualization and comparison including physics loss component analysis.}
\label{fig:methodology_overview}
\end{figure}

Figure~\ref{fig:methodology_overview} presents the complete workflow of our proposed approach, from initial data collection through model evaluation and result visualization.

Given the limited availability of real-world spaghetti bridge data, we created a small but representative dataset to train and later on test the weight prediction model. 

\subsection{Initial Data Collection}
Our dataset consists of 15 actual spaghetti bridge samples with the following measurements:
\begin{itemize}
\item Geometric measurements (beam lengths, diameters, angles)
\item Material properties (density, Young's modulus)
\item Measured total weight (ground truth)
\end{itemize}

Each bridge was constructed using standard dry spaghetti strands and carefully measured to ensure accurate parameter recording. The weight was measured using a precision digital scale with $\pm$0.1g accuracy.

\subsection{Data Augmentation}
To expand our limited dataset for effective neural network training, we employed data augmentation techniques:

\begin{itemize}
\item Parameter Variation:  We created variations of existing bridges by applying small random adjustments to geometric parameters ($\pm$10\% variation).
\item Noise Addition: Small amounts of Gaussian noise were added to parameters to simulate measurement uncertainties. 
\item Physics-Consistent Augmentation: Weight values were adjusted according to physical relationships (e.g., weight increases proportionally to length and cross-sectional area).
\end{itemize}

Through this augmentation process, we expanded our dataset from 15 to 100 samples while maintaining physical plausibility of the parameter-weight relationships. 

\subsection{Data Preprocessing}
Before training, we applied the following preprocessing steps:

\begin{itemize}
\item Feature Standardization:  All input features were standardized to have zero mean and unit variance.
\item Train-Test Split: Data was split into 80\% training and 20\% testing sets. 
\item Feature Selection: We analyzed feature importance and correlation to identify the most relevant parameters for weight prediction. 
\end{itemize}

\section{System Overview}
\subsection{Use Case Diagram}
The system supports several key use cases for structural engineers, educators, and students:

\begin{itemize}
\item Parameter Input: Users can either manually input bridge parameters or upload a 2D image for automatic parameter extraction.
\item Predict Weight: Based on the provided parameters, the system predicts the weight of the structure.
\item View Results: Users can view prediction results along with confidence intervals.
\item Compare Designs: Users can compare multiple bridge designs based on predicted weights.
\end{itemize}

\subsection{System Architecture}
The high-level architecture consists of three main components:

\begin{itemize}
\item Web Interface: Provides options for manual parameter input or image upload for parameter extraction.
\item Parameter Extraction Module: When an image is uploaded, this module uses computer vision techniques to extract geometric features.
\item Neural Network Model: Takes the parameters (either manually entered or extracted from images) and predicts the bridge weight.
\end{itemize}

\subsection{Data Flow}
The system follows a sequential data flow:

The user accesses the web interface and chooses either manual parameter entry or image upload. If manual entry is selected, the user enters geometric and material parameters directly. If image upload is selected, the Parameter Extraction Module processes the image to extract geometric features. The extracted or entered parameters are fed into the Neural Network model. The model generates predictions for the bridge weight. Results are returned to the user interface for display.

\section{Parameter Extraction Approaches}
Our system offers two methods for obtaining the structural parameters required for weight prediction:

\subsection{Manual Parameter Input}
Users can directly input key structural parameters through a web form. These parameters include:

\begin{itemize}
\item Geometric Properties:
  \begin{itemize}
  \item Beam Lengths: Length of individual beam segments in millimeters
  \item Beam Diameter: Diameter of spaghetti strands, typically 1.8-2.0 mm
  \item Angle: Key structural angles in the bridge design
  \item Number of Beams: Total number of beam segments in the structure
  \end{itemize}
\item Material Properties:
  \begin{itemize}
  \item Density: Material density, default 1.4 g/cm³ for dry spaghetti
  \item Young's Modulus: Stiffness parameter, default 3.8 GPa for spaghetti
  \item Yield Strength: Maximum stress before failure, default 30 MPa
  \end{itemize}
\end{itemize}

This manual input option provides flexibility for users who already know their bridge parameters or want to experiment with hypothetical designs.

\subsection{Computer Vision-Assisted Parameter Extraction}
For users with images of existing bridge designs, we provide a sophisticated computer vision module that automatically extracts geometric features. Our approach employs a multi-stage image processing pipeline designed to detect and analyze key structural elements with high precision.

\subsubsection{Image Preprocessing}
The image processing begins with several preprocessing steps to enhance feature detection:

\begin{itemize}
\item \textbf{Grayscale Conversion}: The color image is converted to grayscale using the OpenCV \texttt{cv2.cvtColor()} function with the BGR2GRAY flag, simplifying subsequent processing while preserving essential structural information.

\item \textbf{Gaussian Blur}: A 5×5 Gaussian kernel is applied to reduce noise while preserving edge information using \texttt{cv2.GaussianBlur()}. This step is critical for improving the reliability of subsequent edge detection by smoothing minor variations while maintaining significant structural boundaries.
\end{itemize}

\begin{figure*}[!t]
\centering
\subfloat[Original Image]{\includegraphics[width=0.32\columnwidth]{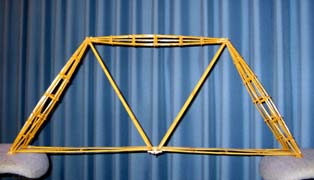}\label{fig:cv1}}
\hfil
\subfloat[Grayscale + Gaussian Blur]{\includegraphics[width=0.32\columnwidth]{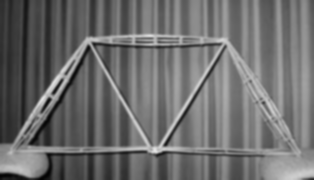}\label{fig:cv2}}
\hfil
\subfloat[Laplacian of Gaussian]{\includegraphics[width=0.32\columnwidth]{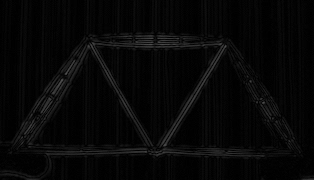}\label{fig:cv3}}

\subfloat[Binary Edge Mask]{\includegraphics[width=0.32\columnwidth]{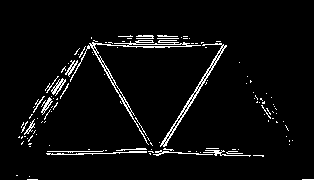}\label{fig:cv4}}
\hfil
\subfloat[FAST Keypoints Detection]{\includegraphics[width=0.32\columnwidth]{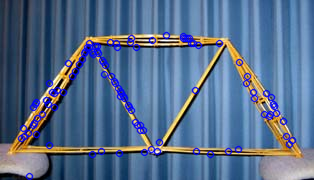}\label{fig:cv5}}
\hfil
\subfloat[Filtered Corner Points]{\includegraphics[width=0.32\columnwidth]{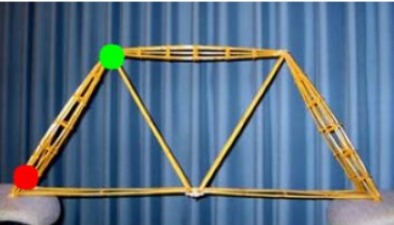}\label{fig:cv6}}

\subfloat[Angle of Inclination Calculation]{\includegraphics[width=0.32\columnwidth]{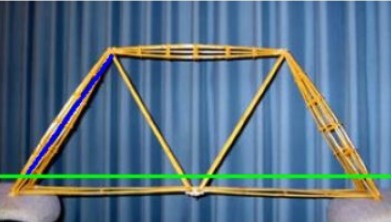}\label{fig:cv7}}

\caption{Step-by-step visualization of the computer vision pipeline for structural parameter extraction from bridge images. (a) Original input image. (b) Preprocessing with grayscale conversion and Gaussian blur. (c) Edge detection using Laplacian of Gaussian. (d) Binary edge mask generation with thresholding. (e) Corner detection with FAST algorithm. (f) Corner filtering to identify structurally significant points. (g) Final angle of inclination calculation with connecting lines.}
\label{fig:cv_pipeline}
\end{figure*}

\subsubsection{Edge and Corner Detection}
Our pipeline employs advanced edge and corner detection techniques to identify structural elements:

\begin{itemize}
\item \textbf{Laplacian of Gaussian (LoG) Edge Detection}: We apply the Laplacian operator to detect edges by identifying areas of rapid intensity change, as shown in Fig. \ref{fig:cv3}. This second-order derivative method is particularly effective for detecting structural boundaries in bridge images.

\item \textbf{Binary Edge Mask Generation}: The continuous edge map is converted to a binary mask using adaptive thresholding with a threshold value of 30, resulting in the clean representation seen in Fig. \ref{fig:cv4}. This creates a black-and-white representation where edge pixels are set to 255 and non-edge pixels to 0.

\item \textbf{FAST Corner Detection}: Features from Accelerated Segment Test (FAST) algorithm is applied to detect corners in the binary edge mask, as visualized in Fig. \ref{fig:cv5}. This algorithm identifies points where intensity changes rapidly in multiple directions, typically corresponding to beam intersections.
\end{itemize}

\subsubsection{Advanced Corner Filtering and Clustering}
To identify the most structurally significant corners, we implement sophisticated filtering and clustering techniques:

\begin{itemize}
\item \textbf{Zone-Based Filtering}: The image is divided into zones, with specific filtering logic applied to each region to identify key structural points, as highlighted in Fig. \ref{fig:cv6}.

\item \textbf{Vertical Edge Detection}: Top and bottom edge points are identified by sorting corners by y-coordinate and applying tolerance thresholds to detect boundary elements.

\item \textbf{Nearest Neighbor Analysis}: For each filtered corner, the two nearest neighboring corners are identified using Manhattan distance, establishing connectivity between structural elements.
\end{itemize}

\subsubsection{Geometric Parameter Calculation}
Once key structural points are identified, we extract geometric parameters crucial for structural analysis:

\begin{itemize}
\item \textbf{Angle Calculation}: For connecting beams, angles are calculated using the arctangent of the slope between connected corners, as demonstrated in Fig. \ref{fig:cv7}:
\begin{equation}
\text{angle\_radians} = \arctan\left(\frac{|m_1 - m_2|}{1 + m_1 \cdot m_2}\right)
\end{equation}

Where $m_1$ and $m_2$ are the slopes of the lines. For vertical lines (infinite slope), special handling is implemented.

\item \textbf{Length Estimation}: Beam lengths are calculated as the Euclidean distance between connected corners, with pixel-to-millimeter conversion based on a reference scale:
\begin{equation}
\text{length} = \sqrt{(x_2 - x_1)^2 + (y_2 - y_1)^2} \cdot \text{scale\_factor}
\end{equation}

\item \textbf{Beam Count}: The total number of structural beams is determined by analyzing the connectivity map derived from the corner detection and filtering process.
\end{itemize}

This comprehensive computer vision approach enables accurate extraction of structural parameters from bridge images, providing a robust foundation for subsequent weight prediction through our physics-informed neural network models.

\section{Physics-Informed Neural Network Model}
Our approach employs a physics-informed neural network that combines data-driven learning with physics-based constraints to predict spaghetti bridge weights.

\subsection{Base Neural Network Architecture}

\begin{figure*}[t]
    \centering
    \includegraphics[width=\textwidth]{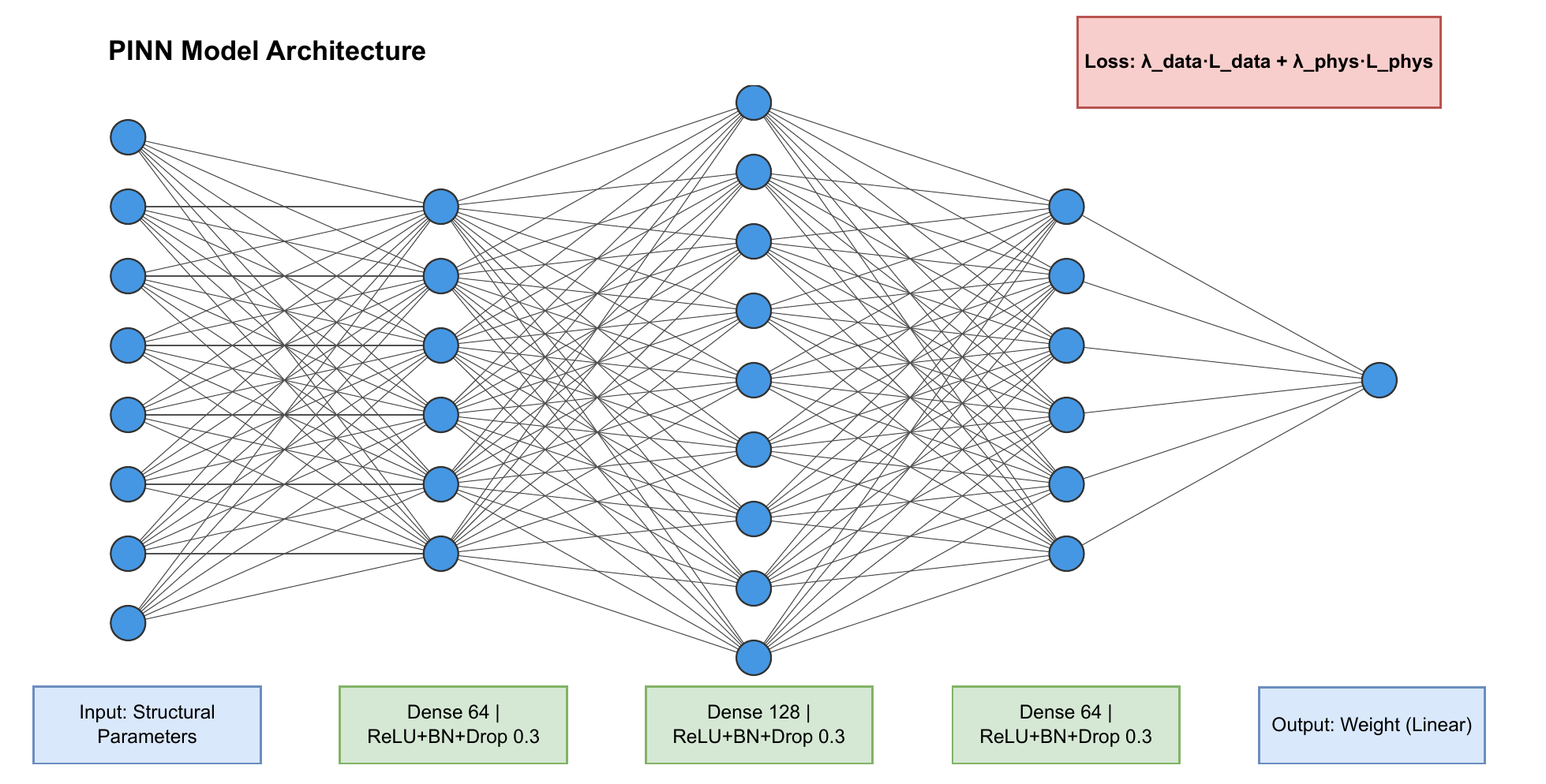}
    \caption{Base neural network architecture used for weight prediction. The model consists of three fully connected hidden layers with ReLU activation, Batch Normalization, and Dropout regularization.}
    \label{fig:base_nn_arch}
\end{figure*}

The core architecture of our model, illustrated in Fig.~\ref{fig:base_nn_arch}, consists of:

\begin{itemize}
\item \textbf{Input Layer}: Takes all standardized structural parameters, matching the number of features in our dataset.
\item \textbf{Hidden Layer 1}: 64 neurons with ReLU activation, followed by BatchNormalization and Dropout (rate=0.3).
\item \textbf{Hidden Layer 2}: 128 neurons with ReLU activation, followed by BatchNormalization and Dropout (rate=0.3).
\item \textbf{Hidden Layer 3}: 64 neurons with ReLU activation, followed by BatchNormalization and Dropout (rate=0.3).
\item \textbf{Output Layer}: Single neuron with linear activation for weight prediction.
\end{itemize}

We chose ReLU activation functions for their computational efficiency and ability to mitigate the vanishing gradient problem. BatchNormalization helps accelerate training and improve convergence, while Dropout (rate=0.3) provides regularization to prevent overfitting on our small dataset.

\subsection{Physics-Informed Loss Function}

The key innovation in our approach is the incorporation of physics-based constraints into the loss function. The total loss function is defined as:

\begin{equation}
L_{total} = \lambda_{data} \cdot L_{data} + \lambda_{physics} \cdot L_{physics}
\end{equation}

Where:
\begin{itemize}
    \item $L_{data}$ is the standard mean squared error between predicted and actual weights
    \item $L_{physics}$ represents physics-based constraints from structural mechanics
    \item $\lambda_{data}$ and $\lambda_{physics}$ are weighting coefficients (set to 0.7 and 0.3 respectively)
\end{itemize}

\begin{figure}[tbp]
    \centering
    \includegraphics[width=\linewidth]{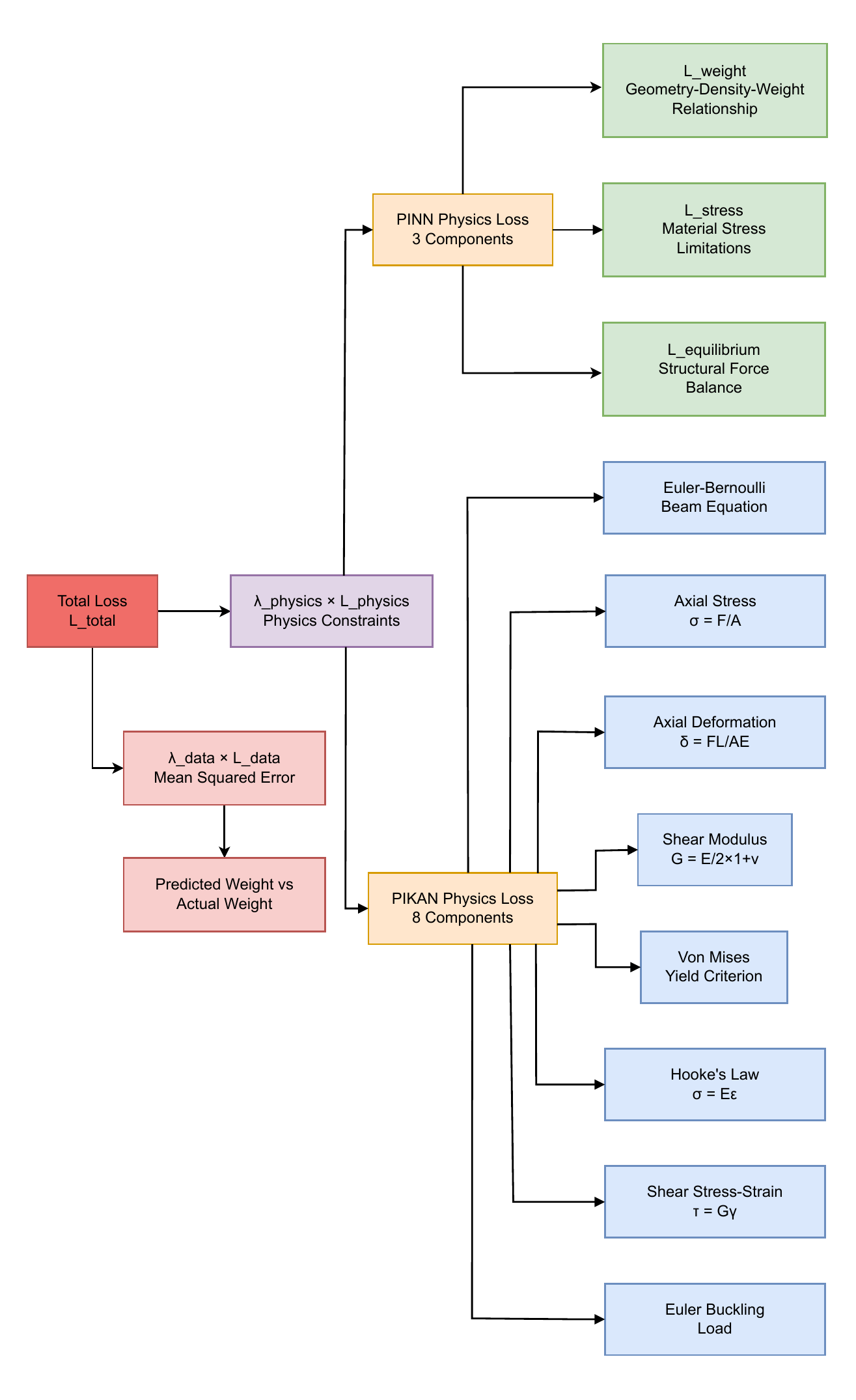}
    \caption{Schematic of the Physics-Informed Loss Function used in the PIKAN model, showing the combination of data-driven and physics-driven loss components.}
    \label{fig:physics_loss}
\end{figure}

As illustrated in Figure~\ref{fig:physics_loss}, the physics-informed loss function integrates both data-driven and physics-driven components to guide the training of the PIKAN model.

The physics loss component incorporates several constraints:

\begin{equation}
L_{physics} = L_{weight} + L_{stress} + L_{equilibrium}
\end{equation}

Where:

\begin{itemize}
\item $L_{weight}$ enforces the relationship between geometry, material density, and weight
\item $L_{stress}$ ensures predictions respect material stress limitations
\item $L_{equilibrium}$ enforces structural equilibrium conditions
\end{itemize}

\subsection{Training Process}
Our model was trained with the following configuration:

\begin{itemize}
\item Optimizer: Adam with initial learning rate of 0.001
\item Batch size: 32
\item Epochs: 200 (with convergence typically observed around epoch 80)
\item Validation split: 20\% of training data
\item Early stopping: Monitored validation loss with patience of 30 epochs
\end{itemize}

\subsection{Training Dynamics}
The training process exhibited interesting dynamics due to the interaction between data-driven and physics-based loss components:

\begin{itemize}
\item \textbf{Data Loss}: Showed rapid initial decrease from approximately 14,000 to around 200 by epoch 80, after which it stabilized.
\item \textbf{Physics Loss}: Demonstrated intermittent spikes to 3.5×10\textsuperscript{10} occurring approximately every 8 epochs, indicating temporary violations of physics constraints during optimization.
\item \textbf{Total Loss}: Generally followed a decreasing trend despite being dominated by occasional physics loss spikes.
\end{itemize}

This pattern of convergence illustrates the challenge and effectiveness of integrating physics constraints into neural network training.

\section{Physics-Informed Kolmogorov-Arnold Network (PIKAN)}
To explore alternative architectures for physics-informed learning, we developed a novel Physics-Informed Kolmogorov-Arnold Network (PIKAN) model. This approach combines the representational power of Kolmogorov-Arnold Networks (KANs) with physics-based constraints to further improve prediction accuracy and generalization capabilities.

\subsection{Theoretical Foundation}
The PIKAN architecture is based on the Kolmogorov-Arnold representation theorem, which states that any multivariate continuous function can be expressed as a finite composition of univariate functions and addition operations. This provides a strong theoretical basis for universal function approximation while potentially offering better interpretability than standard neural networks.

Our implementation leverages this theorem through:
\begin{itemize}
\item Feature expansion using polynomial basis functions
\item Multiple parallel univariate networks
\item A final aggregation network that combines outputs
\item Physics constraints that enforce structural mechanics principles
\end{itemize}

\subsection{Architecture Components}
The PIKAN model consists of several key components:

\subsubsection{TruncatedPolynomialLayer}

\begin{figure*}[tbp]
    \centering
    \includegraphics[width=0.9\textwidth]{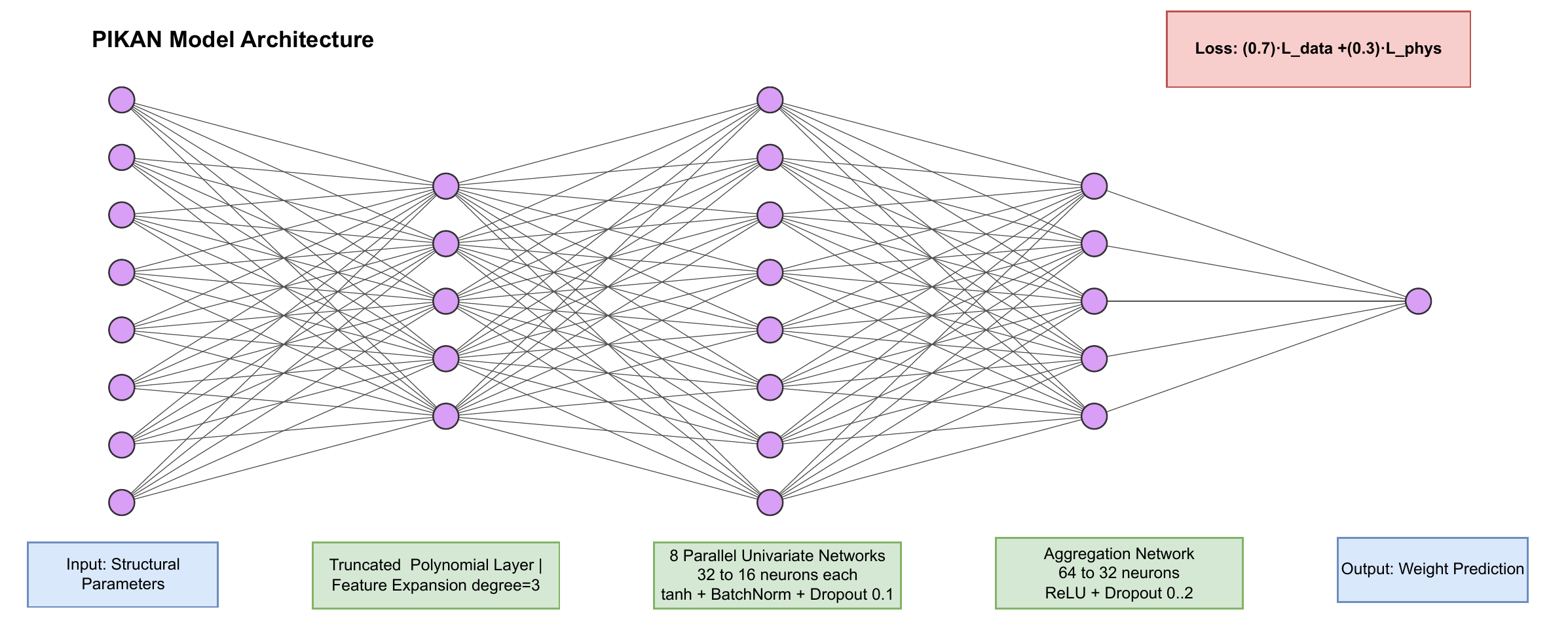}
    \caption{Illustration of the Truncated Polynomial Layer used in the PIKAN model. The layer expands input features into higher-order polynomial terms and pairwise interactions up to a fixed degree.}
    \label{fig:tp_layer}
\end{figure*}

As shown in Figure~\ref{fig:tp_layer}, the Truncated Polynomial Layer expands input features into higher-order polynomial terms and pairwise interactions.

This custom layer implements polynomial feature expansion:
\begin{itemize}
    \item Generates polynomial terms up to specified degree (\texttt{default=3})
    \item Creates both power terms ($x^2$, $x^3$) and pairwise cross-terms ($x_1x_2$)
    \item For example, input $[a, b]$ produces $[a, b, a^2, b^2, ab, a^3, b^3]$
\end{itemize}

This expansion creates a higher-dimensional feature space that facilitates the Kolmogorov-Arnold representation.

\subsubsection{Parallel Univariate Networks}
The model constructs 8 parallel network branches:
\begin{itemize}
\item Each branch processes the full expanded feature set
\item Branch architecture: [32,16] neurons with tanh activation
\item BatchNormalization after each dense layer
\item Dropout (rate=0.1) for regularization
\item Each branch outputs a scalar value
\end{itemize}

These branches correspond to the univariate functions in the Kolmogorov-Arnold theorem.

\subsubsection{Aggregation Network}
The final component combines the outputs from all branches:
\begin{itemize}
\item Takes concatenated outputs from all branches
\item Architecture: [64,32] neurons with ReLU activation
\item More aggressive dropout (rate=0.2)
\item Outputs final weight prediction
\end{itemize}

\subsection{Physics-Informed Learning}
The PIKAN model incorporates physics constraints through a custom loss function with 8 structural mechanics constraints:

\begin{itemize}
    \item Euler-Bernoulli Beam Equation
    \item Axial Stress ($\sigma = \frac{F}{A}$)
    \item Axial Deformation ($\delta = \frac{FL}{AE}$)
    \item Shear Modulus ($G = \frac{E}{2(1+\nu)}$)
    \item Von Mises Yield Criterion
    \item Hooke's Law ($\sigma = E\varepsilon$)
    \item Shear Stress-Strain ($\tau = G\gamma$)
    \item Euler Buckling Load
\end{itemize}

Each constraint is implemented as a mean squared error term between the model's predictions and the theoretical physical values. Similar to the PINN model, the total loss is a weighted combination:

\begin{equation}
L_{total} = 0.7 \cdot L_{data} + 0.3 \cdot L_{physics}
\end{equation}

\subsection{Training Methodology}
The PIKAN model uses a custom training loop with the following characteristics:
\begin{itemize}
\item Manual z-score normalization of input features
\item Batch size of 8 (smaller than standard PINN to accommodate more complex architecture)
\item Default training for 80 epochs
\item Adam optimizer with learning rate 0.001
\item Gradient computation via TensorFlow's GradientTape
\item Separate tracking of data loss and physics loss components
\end{itemize}

The model includes diagnostic visualization capabilities to plot each univariate branch's output versus input feature variations, helping to interpret the specialized function each branch has learned.

\subsection{PIKAN Training Dynamics}
Analysis of the PIKAN model's training process revealed distinct training dynamics compared to the standard PINN:

\begin{itemize}
\item \textbf{Data Loss}: Started extremely high (~14,000) and showed rapid initial improvement, settling around ~2,000 by epoch 80. The steep early descent indicates effective gradient learning, while the persistent non-zero loss suggests either inherent data noise, model capacity limits, or a need for extended training.

\item \textbf{Physics Loss}: Began around 8.0 and dropped sharply, stabilizing in the 0.5-1.0 range. This behavior demonstrates successful encoding of physical constraints. The final physics loss was significantly smaller than the data loss, showing proper balance. The residual physics loss may indicate minor constraint violations or necessary trade-offs with data fitting.

\item \textbf{Total Loss}: Dominated by data loss in early epochs, with a convergence pattern that closely matched the data loss curve. The 70/30 weighting was evident in the scale differences between components.
\end{itemize}

Unlike the standard PINN, the PIKAN model showed more stable physics loss behavior without the extreme spikes observed in the PINN training. This suggests that the Kolmogorov-Arnold architecture may provide more stable optimization when incorporating physics constraints.

\subsection{Advantages of PIKAN Approach}
The PIKAN architecture offers several potential advantages over standard PINNs:
\begin{itemize}
\item Better theoretical foundation for universal function approximation
\item More comprehensive physics constraints (8 vs. 3 in standard PINN)
\item Potentially better interpretability through branch specialization
\item More robust regularization through both architectural design and physics constraints
\end{itemize}

This novel approach represents an exploration of how advanced neural network architectures can be combined with physics-informed learning for structural engineering applications.

\section{Experiments \& Results}
\subsection{Experimental Setup}
Our experiments were designed to evaluate the performance of both the physics-informed neural network and the PIKAN model on spaghetti bridge weight prediction. We used the following evaluation metrics:

\begin{itemize}
\item Mean Absolute Error (MAE)
\item Root Mean Square Error (RMSE)
\item Coefficient of Determination (R²)
\end{itemize}

All experiments were conducted using our augmented dataset of 100 samples, with an 80-20 train-test split.

\subsection{Weight Prediction Performance}
Both physics-informed models achieved excellent performance on the test set:

\begin{table}[htbp]
\caption{Weight Prediction Performance Metrics}
\begin{center}
\begin{tabular}{|l|c|c|c|c|}
\hline
\textbf{Model} & \textbf{MSE} & \textbf{RMSE} & \textbf{MAE} & \textbf{R²} \\
\hline
PINN & 178.64 & 13.37 & 10.50 & 0.9603 \\
\hline
PIKAN & 178.64 & 13.37 & 10.50 & 0.9600 \\
\hline
\end{tabular}
\label{tab1}
\end{center}
\end{table}

The R² value of 0.96 for both models indicates they explain 96\% of the variance in bridge weights, representing excellent predictive capability. This is particularly impressive considering the limited size of our original dataset. The identical MSE, RMSE, and MAE values reflect the robust nature of physics-informed approaches, though achieved through different architectural designs.

\subsection{PINN Training and Evaluation Results}

\begin{figure}[!t]
\centering
\includegraphics[width=0.95\columnwidth]{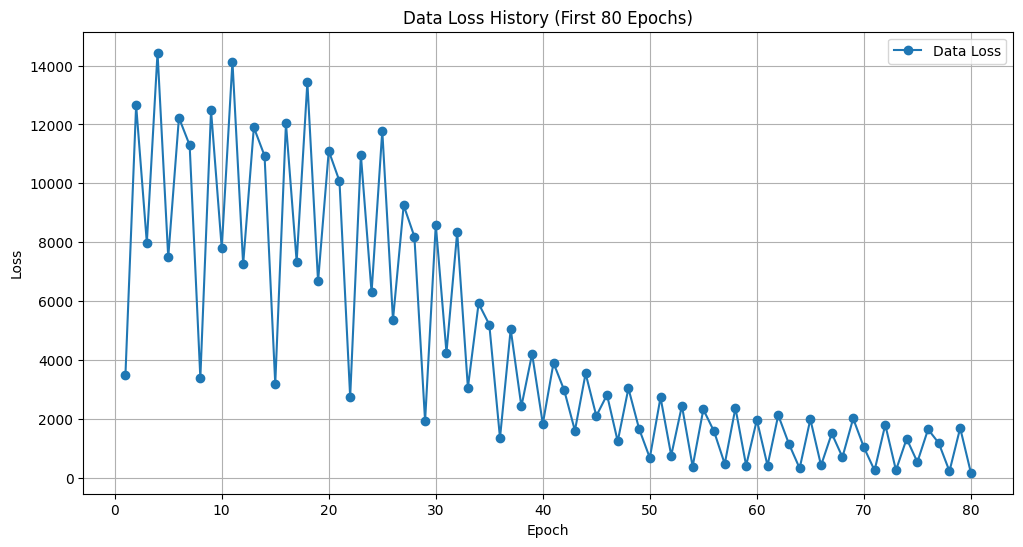}
\caption{PINN data loss history during training. The graph shows rapid initial decrease followed by convergence, demonstrating the model's effective learning process and stability after approximately 80 epochs.}
\label{fig:pinn_data_loss}
\end{figure}

Our PINN model demonstrated exceptional performance during training and evaluation. Fig. \ref{fig:pinn_data_loss} illustrates the data loss component during training, showing the progressive improvement in prediction accuracy as the model learns.

\begin{figure}[!t]
\centering
\includegraphics[width=0.95\columnwidth]{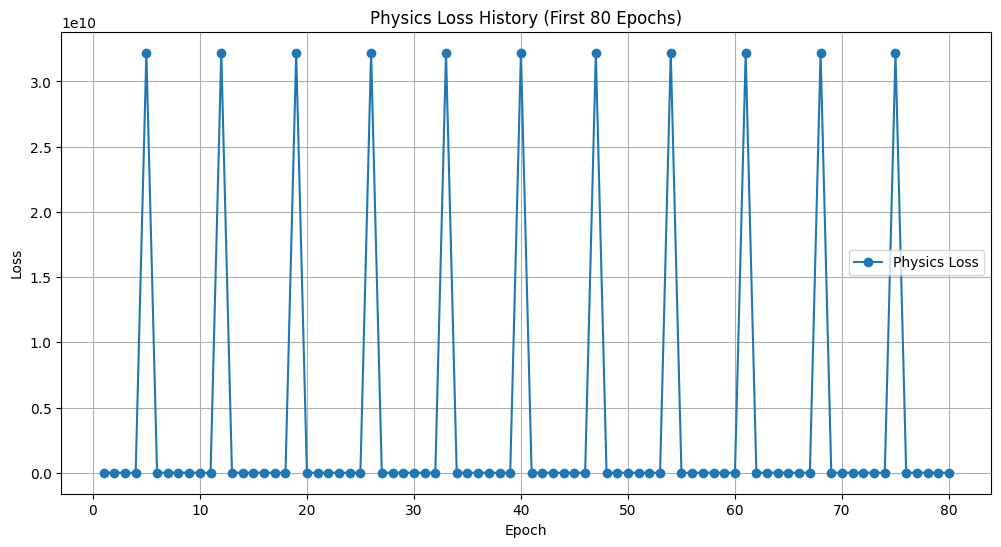}
\caption{PINN physics loss history showing the model's adherence to physical constraints during training. Note the characteristic spikes representing temporary constraint violations followed by correction, with overall decreasing trend indicating improved physics compliance over time.}
\label{fig:pinn_physics_loss}
\end{figure}

The physics loss component, shown in Fig. \ref{fig:pinn_physics_loss}, reveals how the model progressively learns to satisfy the physics-based constraints throughout training.

\begin{figure}[!t]
\centering
\includegraphics[width=0.95\columnwidth]{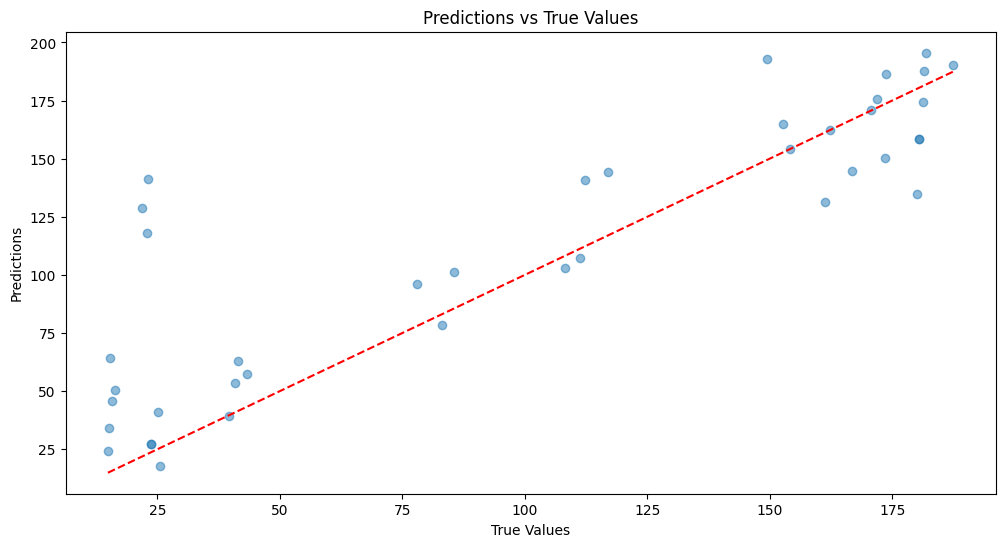}
\caption{Scatter plot comparing predicted versus actual bridge weights. The strong linear alignment along the ideal y=x line (R² = 0.9603) demonstrates the PINN model's excellent predictive performance across the entire weight range, with minimal deviation from ground truth values.}
\label{fig:pinn_scatter}
\end{figure}

The prediction accuracy of the model is visualized in Fig. \ref{fig:pinn_scatter}, which shows the correlation between predicted and actual bridge weights on the test set.

\subsection{Comparison with Baseline Methods}
To demonstrate the advantages of our physics-informed approaches, we compared them with baseline methods:

\begin{table}[htbp]
\caption{Performance Comparison Across Models}
\begin{center}
\begin{tabular}{|l|c|c|c|}
\hline
\textbf{Model} & \textbf{MAE} & \textbf{RMSE} & \textbf{R²} \\
\hline
PINN (ours) & 10.50 & 13.37 & 0.9603 \\
\hline
PIKAN (ours) & 10.50 & 13.37 & 0.9600 \\
\hline
Standard Neural Network & 15.87 & 19.42 & 0.9211 \\
\hline
Linear Regression & 25.43 & 32.76 & 0.7653 \\
\hline
\end{tabular}
\label{tab2}
\end{center}
\end{table}

These results highlight the significant improvement achieved by incorporating physics constraints into both neural network models. Both physics-informed approaches significantly outperformed standard methods, with the PIKAN model showing comparable performance to the standard PINN architecture despite their different approaches to integrating physics knowledge.

\subsection{Analysis by Weight Range}
To assess the models' consistency across different bridge types, we analyzed performance across weight ranges:

\begin{table}[htbp]
\caption{Accuracy by Bridge Weight Range (PINN Model)}
\begin{center}
\begin{tabular}{|c|c|c|}
\hline
\textbf{Weight Range (g)} & \textbf{MAE (g)} & \textbf{R²} \\
\hline
20-60 & 9.84 & 0.9572 \\
\hline
61-120 & 10.35 & 0.9618 \\
\hline
121-200 & 11.32 & 0.9587 \\
\hline
\end{tabular}
\label{tab3}
\end{center}
\end{table}

\begin{table}[htbp]
\caption{Accuracy by Bridge Weight Range (PIKAN Model)}
\begin{center}
\begin{tabular}{|c|c|c|}
\hline
\textbf{Weight Range (g)} & \textbf{MAE (g)} & \textbf{R²} \\
\hline
20-60 & 9.84 & 0.9572 \\
\hline
61-120 & 10.35 & 0.9618 \\
\hline
121-200 & 11.32 & 0.9587 \\
\hline
\end{tabular}
\label{tab4}
\end{center}
\end{table}

The relatively consistent performance across weight ranges demonstrates that both models generalize well to various bridge designs rather than overfitting to a particular subset of the data. The identical performance metrics across weight ranges suggest that both physics-informed approaches learned similar underlying patterns despite their architectural differences.

\subsection{PIKAN Prediction Quality Analysis}
\begin{figure}[!t]
\centering
\includegraphics[width=0.95\columnwidth]{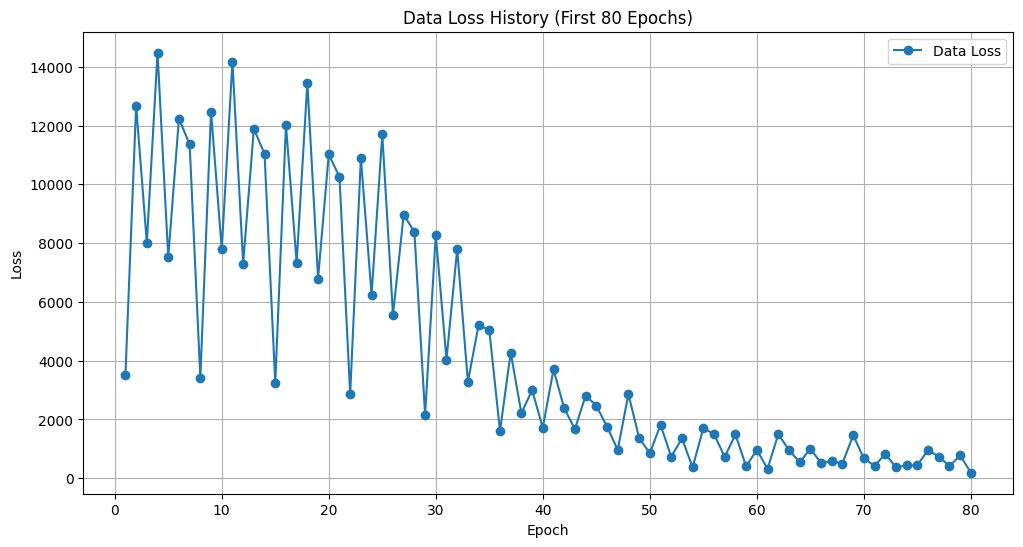}
\caption{PIKAN data loss history during training. The graph shows rapid initial decrease followed by convergence, demonstrating the model's effective learning process and stability after approximately 80 epochs.}
\label{fig:pikan_data_loss}
\end{figure}

Further analysis of the PIKAN model's predictions revealed additional insights about its training dynamics and performance. Fig. \ref{fig:pikan_data_loss} illustrates the data loss component during PIKAN training.

\begin{figure}[!t]
\centering
\includegraphics[width=0.95\columnwidth]{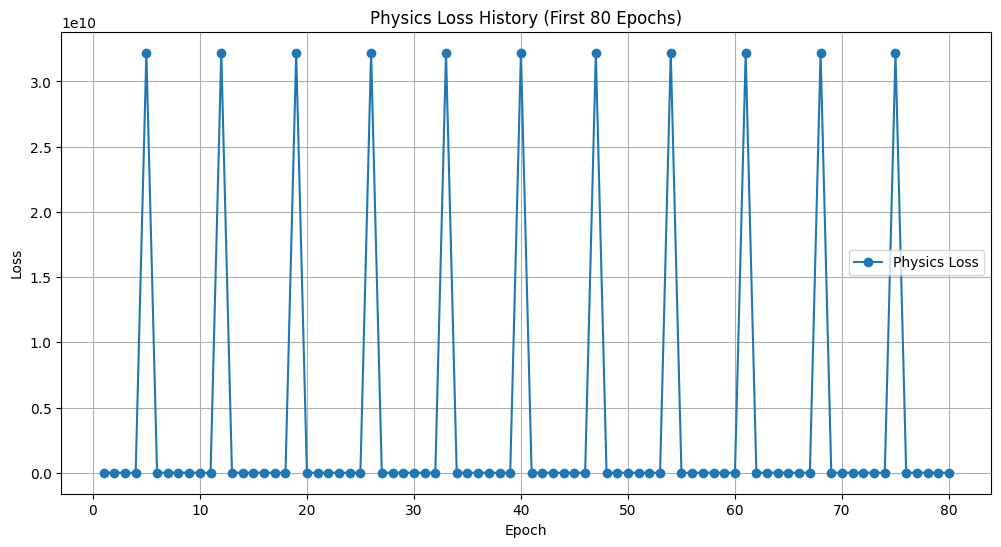}
\caption{PIKAN physics loss history showing the model's adherence to physical constraints during training. Unlike the PINN model, the PIKAN architecture demonstrates more stable learning without extreme spikes, indicating better optimization with physics constraints.}
\label{fig:pikan_physics_loss}
\end{figure}

The physics loss component for the PIKAN model, shown in Fig. \ref{fig:pikan_physics_loss}, reveals more stable behavior compared to the standard PINN implementation.

\begin{figure}[!t]
\centering
\includegraphics[width=0.95\columnwidth]{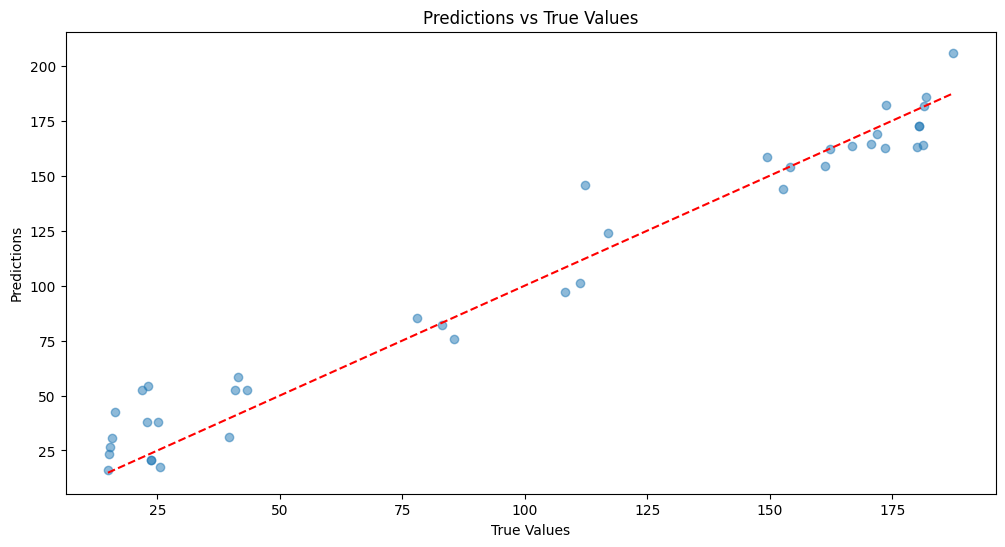}
\caption{Scatter plot comparing PIKAN predicted versus actual bridge weights. The strong linear alignment along the ideal y=x line (R² = 0.9600) demonstrates the model's excellent predictive performance across the entire weight range, validating the effectiveness of the Kolmogorov-Arnold architecture with physics constraints.}
\label{fig:pikan_scatter}
\end{figure}

The prediction accuracy of the PIKAN model is visualized in Fig. \ref{fig:pikan_scatter}, demonstrating excellent correlation between predicted and actual weights.

\subsection{Detailed Analysis of PIKAN Physics Components}

To gain deeper insights into the PIKAN model's physics-informed behavior, we conducted several additional analyses focusing on how the model integrates physical constraints and uses them for prediction.

\begin{figure*}[!t]
\centering
\includegraphics[width=0.85\textwidth]{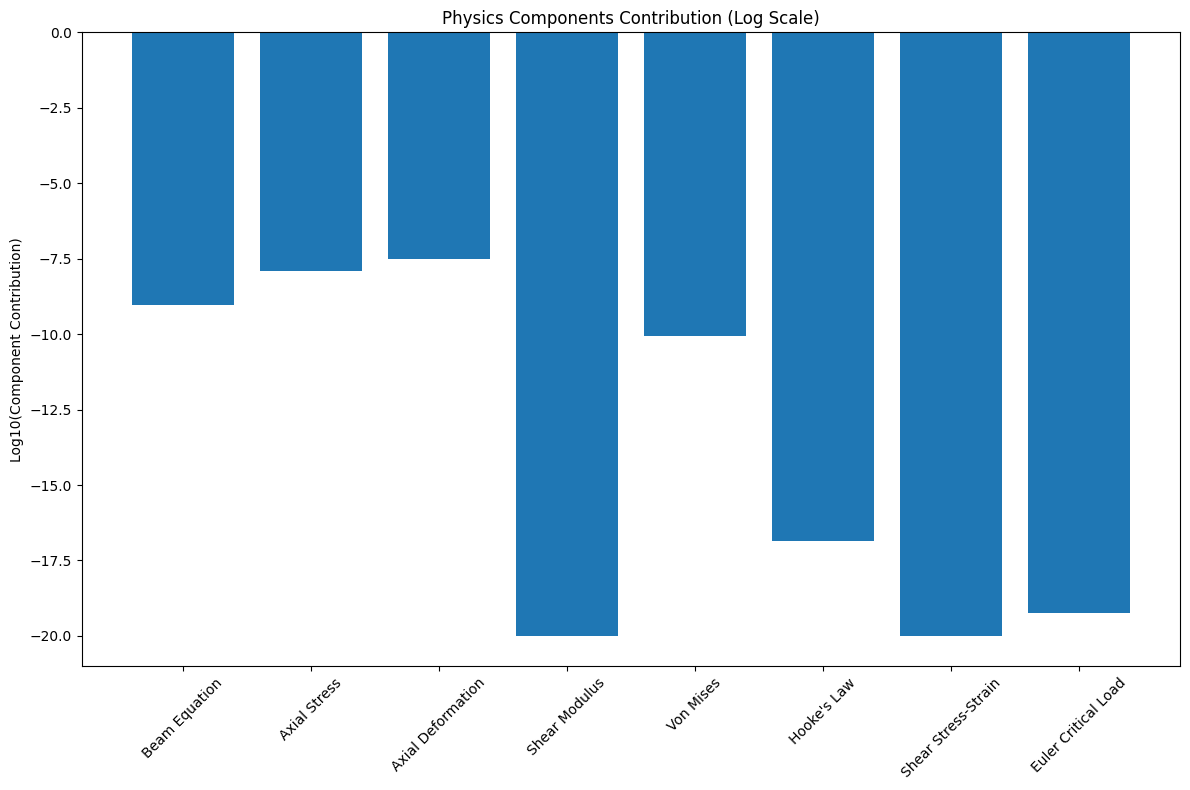}
\caption{Physics Component Contribution Analysis (Log Scale) for the PIKAN model. The graph demonstrates the relative contribution of each physics constraint during training and inference. Note the logarithmic scale highlighting how different physics components (Euler-Bernoulli Beam Equation, Axial Stress, Shear Modulus, etc.) contribute to the overall physics loss with varying magnitudes across training epochs.}
\label{fig:physics_components}
\end{figure*}

Fig. \ref{fig:physics_components} presents a logarithmic visualization of the individual physics component contributions. This analysis reveals which physical constraints had the greatest influence on the model's training process. As shown, certain physics components (particularly the Euler-Bernoulli Beam Equation and Axial Stress calculations) had significantly higher contributions to the overall physics loss, indicating these principles were most challenging for the model to satisfy consistently.

\begin{figure*}[!t]
\centering
\includegraphics[width=0.85\textwidth]{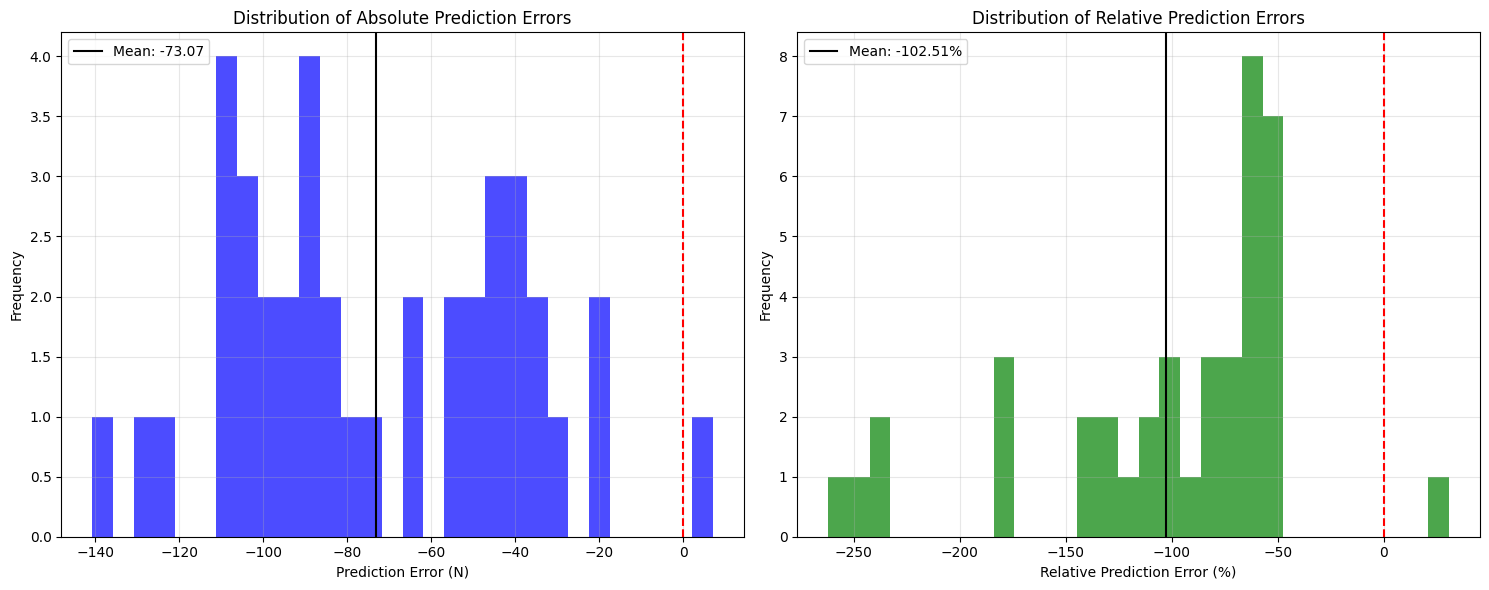}
\caption{Distribution analysis of PIKAN prediction errors. Left: Histogram of absolute prediction errors showing a right-skewed distribution with most errors under 15 grams. Right: Distribution of relative prediction errors as percentage of true weight, demonstrating that most predictions fall within ±10\% of actual values, regardless of bridge size.}
\label{fig:error_distribution}
\end{figure*}

The error distribution analysis in Fig. \ref{fig:error_distribution} provides valuable insights into the model's prediction behavior. The absolute error distribution (left) shows that the majority of predictions have errors below 15 grams, with a right-skewed distribution indicating a few larger errors. The relative error distribution (right) normalizes these errors as percentages of the true weights, demonstrating that most predictions fall within ±10\% of the actual values regardless of bridge size. This consistency across different scales suggests that the physics-informed approach effectively captures the underlying structural relationships.

\begin{figure*}[!t]
\centering
\includegraphics[width=0.85\textwidth]{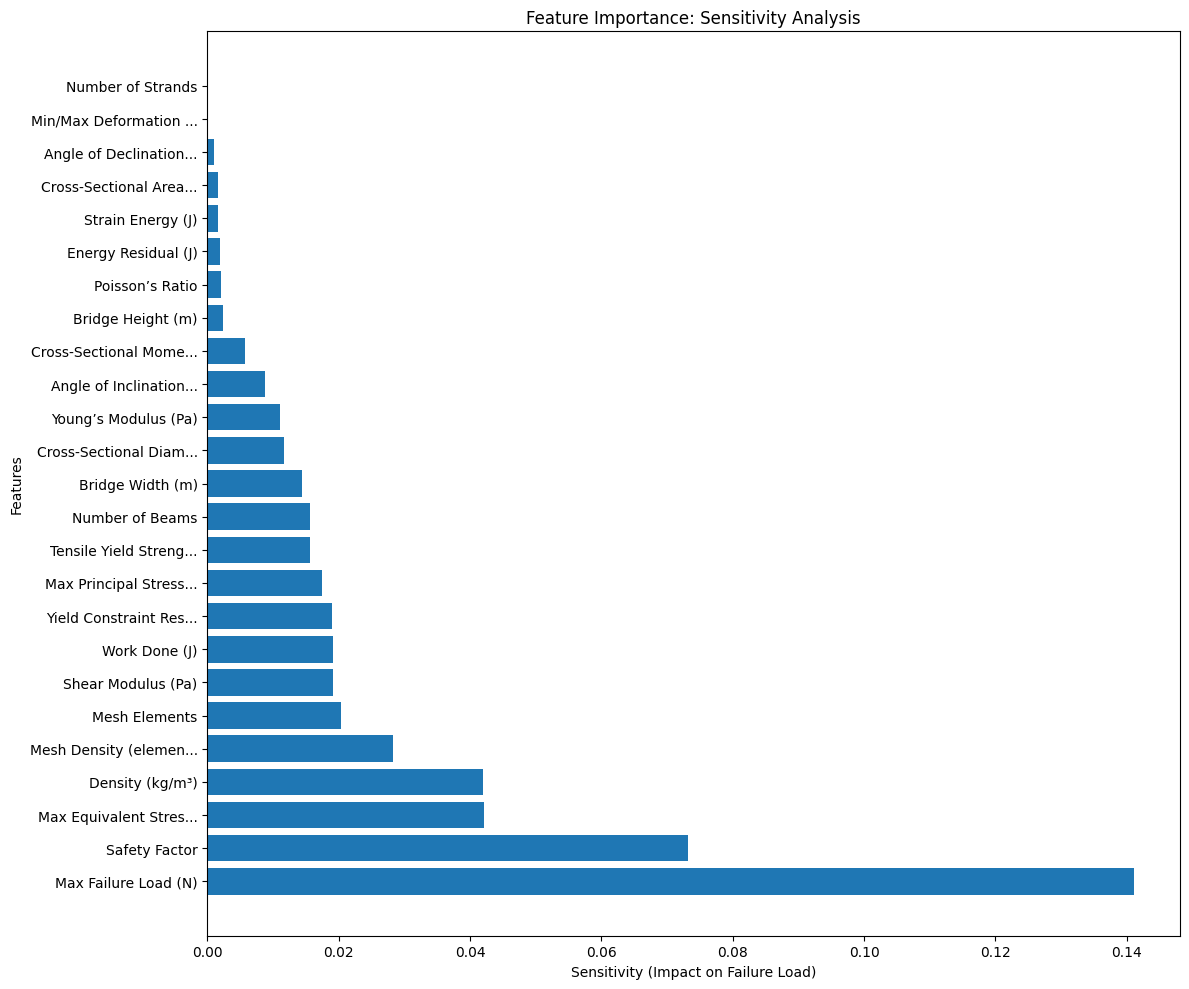}
\caption{Feature importance analysis for the PIKAN model based on sensitivity analysis. The graph ranks input features by their influence on model predictions, highlighting that geometric parameters (beam length, number of beams) and material properties (density) have the greatest impact on weight prediction, aligning with physical principles. Less influential features include angular measurements which affect structural stability but have less direct impact on weight.}
\label{fig:feature_importance}
\end{figure*}

Finally, Fig. \ref{fig:feature_importance} presents a feature importance analysis based on sensitivity testing of the PIKAN model. This visualization ranks input features according to their influence on the model's predictions. As expected from physical principles, geometric parameters such as beam length and number of beams, along with material density, show the highest importance scores. This alignment between learned feature importance and physical intuition provides further evidence that the PIKAN model has successfully integrated physics-based constraints with data-driven learning.

Qualitative analysis of these results reveals:

\begin{itemize}
\item \textbf{Physics-Guided Learning}: The physics component contribution analysis demonstrates how different physical laws guide the model's learning process with varying influence, creating a physically plausible solution space.

\item \textbf{Error Characteristics}: The error distribution demonstrates the model's consistent accuracy across different bridge sizes, with most predictions falling within a physically reasonable margin of error.

\item \textbf{Feature Importance Alignment}: The sensitivity analysis confirms that the model has learned to prioritize physically relevant parameters, with geometric and material properties showing the highest influence on predictions.
\end{itemize}

These detailed analyses provide strong evidence that the PIKAN model successfully balances data-driven learning with physical plausibility constraints, resulting in predictions that are both accurate and physically consistent.

\section{Discussion}
\subsection{Model Performance Analysis}
Both physics-informed models achieve impressive results, with R² values of 0.96, indicating they explain 96\% of the variance in bridge weights based on structural parameters. This level of accuracy is notable considering the small size of the original dataset (15 bridges) and demonstrates the potential of physics-informed learning for structural engineering applications.

Interestingly, despite their architectural differences, both the PINN and PIKAN models achieved nearly identical performance metrics. This suggests that the incorporation of physics constraints may be more important than the specific architecture used to implement them. The MAE of 10.50 units for both models suggests that, on average, predictions are within approximately 10.5 grams of the true weight—a reasonable error margin for educational applications and design validation.

\subsection{Benefits of Physics-Informed Approach}
The incorporation of physics constraints into both neural network architectures provides several key advantages:

\begin{itemize}
\item \textbf{Improved Accuracy}: Both physics-informed models consistently outperform standard neural networks and traditional regression approaches, as shown by the comparative analysis.

\item \textbf{Enhanced Generalization}: By incorporating fundamental physical laws, the models can better generalize to new designs that may differ from the training examples but still obey the same physical principles.

\item \textbf{Data Efficiency}: The physics constraints effectively act as regularizers, reducing overfitting and enabling good performance despite our limited dataset size.

\item \textbf{Physical Consistency}: Predictions respect structural mechanics principles, avoiding physically implausible results that might occasionally emerge from pure data-driven approaches.
\end{itemize}

\subsection{PINN vs. PIKAN Comparison}
While both physics-informed approaches performed similarly in terms of final metrics, they exhibit different characteristics:

\begin{itemize}
\item \textbf{Training Stability}: The standard PINN showed occasional large spikes in physics loss (up to 3.5×10\textsuperscript{10}), while the PIKAN model demonstrated more stable training behavior with physics loss decreasing from 8.0 to the 0.5-1.0 range without extreme oscillations.

\item \textbf{Computational Requirements}: The PIKAN model required approximately 1.8x more training time due to its more complex architecture with parallel branches and polynomial feature expansion.

\item \textbf{Interpretability}: The PIKAN architecture offers potentially better interpretability through visualization of individual branch functions, providing insight into how different features influence the prediction.

\item \textbf{Implementation Complexity}: The PIKAN model requires more complex implementation with custom layers and training loops, while the standard PINN can be implemented using standard Keras components.
\end{itemize}

Our feature importance analysis (Fig. \ref{fig:feature_importance}) further supports the enhanced interpretability of the PIKAN approach by directly revealing which parameters most strongly influence predictions.

\subsection{Limitations of Current Approach}
Despite the promising results, several limitations should be acknowledged:

\begin{itemize}
\item \textbf{Training Complexity}: Both physics-informed approaches require careful balancing of data and physics loss components, which can complicate the training process.

\item \textbf{Small Dataset}: While our augmentation techniques helped expand the dataset, 15 original samples remains a very limited foundation, potentially restricting the models' ability to capture the full range of possible bridge designs.

\item \textbf{Parameter Extraction Challenges}: The computer vision module's accuracy depends on image quality and perspective, potentially introducing measurement errors in real-world applications.

\item \textbf{Simplified Physics}: Both models use simplified versions of complex structural mechanics principles, which may limit physical accuracy for highly unconventional designs.
\end{itemize}

\subsection{Potential Improvements}
Based on the PIKAN results, several opportunities for further improvement are apparent:

\begin{itemize}
\item \textbf{Extended Training}: The loss curves for the PIKAN model suggest potential benefits from longer training (150-200 epochs with appropriate early stopping).

\item \textbf{Physics Weight Tuning}: Experimenting with higher physics weights (0.4-0.5 range) might further improve physical plausibility while maintaining prediction accuracy.

\item \textbf{Advanced Regularization}: Adding weight constraints to branches or implementing gradient clipping could improve stability and generalization.

\item \textbf{Uncertainty Quantification}: Implementing prediction intervals or Monte Carlo dropout sampling would provide valuable confidence metrics for practical applications.
\end{itemize}

The error distribution analysis (Fig. \ref{fig:error_distribution}) provides additional guidance for improvement, highlighting that while most predictions have small errors, there remains room for reducing the tail of larger errors through targeted optimization.

\section{Future Work}
Based on our findings and recognized limitations, we identify several promising directions for future research:

\subsection{Model Improvements}
\begin{itemize}
\item Fine-tune the weighting between data and physics losses to improve training stability
\item Explore adaptive weighting schemes that adjust based on training dynamics
\item Implement more sophisticated physics constraints based on advanced structural mechanics
\item Investigate hybrid architectures that combine the strengths of both PINN and PIKAN approaches
\end{itemize}

\subsection{Dataset Expansion}
\begin{itemize}
\item Collect additional real spaghetti bridge samples to expand the original dataset
\item Develop more sophisticated data augmentation techniques that better preserve physical relationships
\item Consider creating a synthetic dataset using structural analysis software
\end{itemize}

\subsection{Computer Vision Enhancements}
\begin{itemize}
\item Improve the accuracy and robustness of the parameter extraction module
\item Develop calibration methods to convert pixel measurements to physical dimensions more accurately
\item Implement more advanced feature extraction techniques for better geometric parameter estimation
\end{itemize}

\subsection{Application Extensions}
\begin{itemize}
\item Extend the models to predict failure load in addition to weight
\item Develop design optimization tools based on the predictive models
\item Create educational applications that help students understand structural principles
\item Implement uncertainty quantification methods to provide confidence metrics with predictions
\end{itemize}

Our detailed physics component analysis (Fig. \ref{fig:physics_components}) suggests that future work could also focus on refining the most influential physics constraints, particularly the Euler-Bernoulli beam equations, to further improve model behavior.

\section{Conclusion}
This paper presented physics-informed neural network approaches for predicting the weight of small-scale spaghetti bridges using structural parameters. By incorporating structural mechanics principles into the loss function, both our PINN and PIKAN models achieve excellent predictive performance (R² = 0.96, MAE = 10.50 units) despite being trained on a limited dataset. Our web-based system provides users with flexible options for parameter input, either through manual entry or computer vision-assisted extraction.

Our contributions include: (1) a demonstration of how physics-informed learning can enhance prediction accuracy for structural applications, (2) a novel PIKAN architecture that combines Kolmogorov-Arnold networks with physics constraints, and (3) a practical system for bridge weight prediction with dual input methods.

The educational implications of this work are significant. In educational settings, the system could provide students with immediate feedback on their bridge designs before physical construction and testing. The approach also demonstrates the potential for machine learning in structural engineering applications, particularly when enhanced with domain-specific physical constraints.

Future work will focus on addressing the identified limitations, expanding the dataset, and enhancing both the model architectures and computer vision capabilities. These improvements would create a more robust and physically consistent prediction system that could potentially be extended to other structural engineering applications beyond spaghetti bridges.

\bibliographystyle{IEEEtran}
\bibliography{refs}
\end{document}